\documentclass[journal,twocolumn]{IEEEtran}
\usepackage{xcolor}
\usepackage{booktabs}
\usepackage{mathtools}
\usepackage{pifont}
\usepackage{epsfig,makeidx,color,subfigure}
\usepackage{amsmath,amssymb,bbm,enumitem}
\usepackage{cite,graphicx,lipsum}
\usepackage[switch,pagewise]{lineno}
\usepackage{tabularx}
\usepackage{hyperref}
\hypersetup{
        colorlinks = true,
        citecolor = red,
        breaklinks = true,
        linktocpage=true,
}
\usepackage{url}

\def\be{ \begin{equation} }
\def\ee{ \end{equation} }
\def\bea{ \begin{eqnarray} }
\def\eea{ \end{eqnarray} }

\def\bx{{\bf x}}

\def\bz{{\bf z}}

\def\b0{{\bf 0}}

\def\cD{{\cal D}}

\def\cZ{{\cal Z}}

\ifCLASSOPTIONonecolumn
\else

\fi

\begin{document}

\title{Generative AI Meets Semantic Communication: Evolution and Revolution of Communication Tasks}

\author{Eleonora Grassucci,~\IEEEmembership{Member,~IEEE}, Jihong Park,~\IEEEmembership{Senior Member,~IEEE}, Sergio Barbarossa,~\IEEEmembership{Fellow,~IEEE}, Seong-Lyun Kim,~\IEEEmembership{Member,~IEEE}, Jinho Choi,~\IEEEmembership{Fellow,~IEEE}, Danilo Comminiello,~\IEEEmembership{Senior Member,~IEEE}\thanks{E. Grassucci, S. Barbarossa, and D. Comminiello are with the Department of Information Engineering, Electronics and Telecommunication of Sapienza University of Rome, Italy. J. Park and J. Choi are with the School of Information Technology, Deakin University, Geelong, VIC 3220, Australia. S.-L. Kim is with the School of Electrical and
Electronic Engineering, Yonsei University, Seoul, South Korea.}}

\maketitle
\begin{abstract}
While deep generative models are showing exciting abilities in computer vision and natural language processing, their adoption in communication frameworks is still far underestimated. These methods are demonstrated to evolve solutions to classic communication problems such as denoising, restoration, or compression. Nevertheless, generative models can unveil their real potential in semantic communication frameworks, in which the receiver is not asked to {\it recover} the sequence of bits used to encode the transmitted (semantic) message, but only to {regenerate} content that is {\it semantically consistent} with the transmitted message. 
Disclosing generative models capabilities in semantic communication paves the way for a paradigm shift with respect to conventional communication systems, which has great potential to reduce the amount of data traffic and offers a revolutionary versatility to novel tasks and applications that were not even conceivable a few years ago. In this paper, we present a unified perspective of deep generative models in semantic communication and we unveil their revolutionary role in future communication frameworks, enabling emerging applications and tasks. Finally, we analyze the challenges and opportunities to face to develop generative models specifically tailored for communication systems.
\end{abstract}

\begin{IEEEkeywords}
Generative Semantic Communication, Generative Models, Semantic Communication, Emerging Applications
\end{IEEEkeywords}

\ifCLASSOPTIONonecolumn
\baselineskip 28pt
\fi

\section{Introduction}
\label{sec:intro}

In the last few months, deep generative models such as Stable Diffusion for image and ChatGPT for conversational text generation forcefully entered our daily lives, and it has immediately become clear that such models will revolutionize many different fields of research and productivity. The generative process behind these models starts by assuming that the initial set of data obeys a probability distribution that has to be inferred from data. During training, deep generative models learn to estimate such distribution to be able to generate new samples from it at inference time. This generation process can be guided or conditioned to obtain the desired generated output (so that, for instance, the prompt ``a dog running in the backyard" leads to the generated of an image containing a dog running over a background). Here, we focus on the role of generative deep learning in 6G communications and how it may open new perspectives and opportunities.


\begin{figure}[t]
    \centering
    \includegraphics[width=\linewidth]{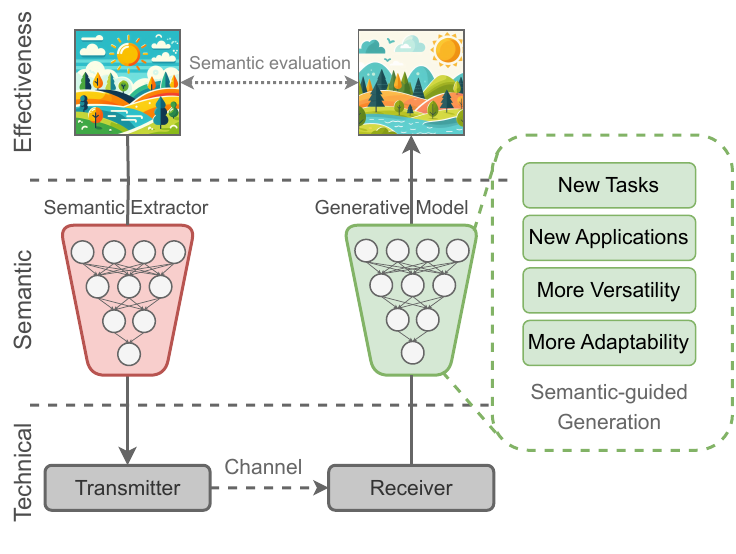}
    \caption{The three levels of Weaver model with generative models that lie in the semantic level enabling new tasks and applications under the semantic communication paradigm. The generated output will then be evaluated under semantic metrics.}
    \label{fig:levels}
\end{figure}

Historically speaking, in his article accompanying Shannon's Mathematical Theory of Communication, Weaver identified three levels of communication \cite{shannon1949}, 
as schematically depicted in Fig. \ref{fig:levels}. The top level, the so-called effectiveness level, identifies the goal of the communication. This level operates under specific requirements on key performance indicators that measure how well the (semantic) message conveyed by the source allows the destination to fulfill the task. The middle level, the semantic level, includes a semantic extractor, at the transmitter side, that extracts the features necessary at the receiver side to regenerate a message that is semantically equivalent to the transmitted one. The bottom level represents the technical level, where the relevant information is encoded using conventional approaches, but that are aware of the higher levels. Shannon deliberately focused on the technical level, concerned with the encoding of the source message into a sequence of bits and the recovery of this sequence after passing through a noisy channel, independently of the meaning conveyed by the message.
Very recently, a new paradigm has emerged for wireless communications, where the attention is progressively moving to the semantic level, devising the so-called semantic communication. This field is built upon the idea that what matters is to regenerate, at the receiver side, an information that is {\it semantically equivalent} to the transmitted one without necessarily coinciding, at the bit (symbolic) level, with the sequence of transmitted bits \cite{strinati20216g}, \cite{barbarossa2023semantic}.
This idea offers many additional degrees of freedom that can be advantageously exploited in the design of the system and in the resource allocation \cite{Dai2021CommunicationBT}. In particular, this novel paradigm may unveil the big potential of deep generative models by using them to regenerate content at the receiver side, conditioned to only a very compact representation of the transmitted information.

Generative semantic communication is the melting point between generative deep learning and semantic communications \cite{barbarossa2023semantic}\cite{Grassucci2023GenerativeSC}. Generative models are indeed extremely versatile in solving multiple tasks thus enabling the goal-oriented communication paradigms \cite{strinati20216g}.
In addition, thanks to the unsupervised nature of deep generative models, generative semantic communication frameworks do not need to be trained end-to-end, enabling more flexible frameworks. 
Moreover, the real breakthrough, still unveiling, that deep generative models are bringing to communication frameworks paves the way for completely new tasks and applications that were not even conceivable years ago, such as multi-user or personalized communication, content creation, multimodal generation, or language-oriented communication.

Together with exciting results, deep generative models such as diffusion or large language models come with new challenges. To puzzle out their large demand for computational resources during both training and inference, captivating solutions have been unfolding and will be further developed in the near future towards the evolution of generative semantic communication frameworks.

The contributions of this paper are threefold and can be summarized as follows.

\begin{itemize}
    \item We propose a unified perspective of deep generative models in semantic communications and of their transformative role in the future sixth generation.
    \item We unveil the revolution these models bring to semantic communication, enabling new tasks and unleashing completely novel applications that go beyond classical communication problems.
    \item We suggest some solutions to generative modeling challenges on high computational resources demanding, running towards novel and efficient generative semantic communication frameworks.
\end{itemize}


\section{Generative Models for Semantic Communication}

\begin{figure*}
    \centering
    \includegraphics[width=\textwidth]{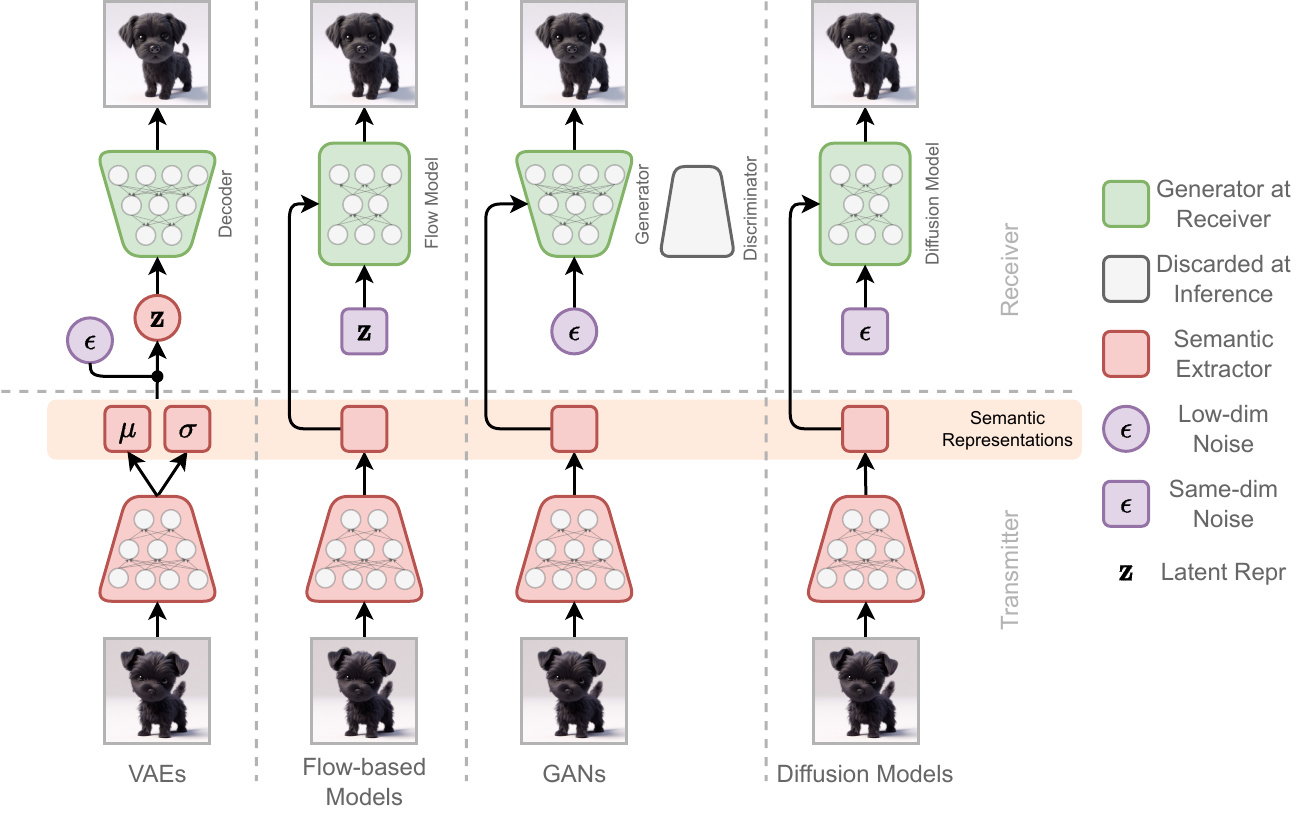}
    \caption{Taxonomy of deep generative models for semantic communication.}
    \label{fig:taxonomy}
\end{figure*}

\subsection{Generative Model Architectures}
\label{sec:taxonomy}

At the receiver side of the Weaver model in Fig.~\ref{fig:levels}, the semantic level involves a deep generative model conditioned to the received semantics to regenerate content and unlock completely new applications. 
This subsection presents the most common methods among the wide range of deep generative models, their advantages, and the limitations of their effective development in communication systems. A taxonomy of generative semantic communication frameworks with the most common generative models is shown in Fig.~\ref{fig:taxonomy}.


\textbf{Variational Autoencoders.} Variational Autoencoders (VAEs) 
comprise two neural networks: the first one encodes the mean and variance of the data that are then involved in adjusting a Gaussian distribution to build a lower-dimensional latent representation of the data.
During transmission, we can transmit the extracted statistics and perform the reparameterization with the Gaussian distribution at the receiver. This allows great flexibility for the receiver in the case of bad channel conditions in which part of the transmitted content may be lost. Indeed, during training the latent representation is forced to be as close as possible to a standard Gaussian distribution and, at inference time, if a portion of the transmitted information is lost, it is possible to infer the learned statistics to preserve the ideal distribution of the latent representation \cite{barbarossa2023semantic}. The second network, the decoder, can then take in input the latent vector and regenerate the content according to the transmitted features. Therefore, in VAEs the encoder acts as a semantic extractor at the transmitter and the decoder as a generator at the receiver, in a stable end-to-end (E2E) training fashion. VAEs can provide lossy compression and their sampling is fast, therefore useful in practical applications.


\textbf{Flow-based Models.} Flow-based models are based on a sequence of invertible transformations that starts with a simple distribution and turns it into a complex one (e.g., the data one). Differently from VAEs, they usually do not have an encoder for learning semantic representations, therefore, to employ flow models in semantic communication frameworks, we should insert a semantic extractor at the transmitter, whose output will condition the flow process to generate content with consistent semantics. 
Flow-based models can provide lossless compression and their training is very stable, but they suffer from slow sampling and sometimes limited expressiveness.

\textbf{Generative Adversarial Networks.} Generative Adversarial Networks (GANs) 
are composed of two networks, a generator that takes in input a noise vector and outputs the generated content, and a discriminator, whose aim is to distinguish generated from real content and provide feedback to the generator to improve the sample quality. At inference time, one can discard the discriminator and involve only the generator to generate samples. As for Flow models, in a communication scenario, we have to condition the generation with some received information. To this end, we can build a semantic communication framework by inserting a semantic extractor in the sender and transmitting the compressed semantics to the receiver is involved in guiding the generation process of the generator. The semantic extractor and the generator do not need end-to-end training, increasing the framework flexibility. Furthermore, GANs have usually higher expressiveness than VAEs and Flows in generation abilities, and their sampling process is very fast, although their training is highly unstable and hyperparameter-dependent.

\textbf{Diffusion Models.}
Denoising Diffusion Probabilistic Models (DDPMs) have recently gained the floor among deep generative models due to their exciting results in dealing with high-quality, multimedia, and multimodal data, beating all the previous competitors. They have been recently introduced also in semantic communication \cite{Grassucci2023GenerativeSC, Grassucci2023DiffusionMF, pezone2023semantic}.
Diffusion models are based on a Gaussian Markov chain that progressively transforms the content into pure noise, training a neural network to estimate the amount of added noise. The idea behind this approach is that if the model learns this direction of the transformation from data to noise, then once trained it will be able to perform the reverse process. Therefore, at inference time, starting from a sample of pure noise the network will progressively remove noise up to delineate a newly generated sample. As for GANs, the generation process is unconditional and, in a communication scenario, we are required to condition the process with the transmitted information \cite{Grassucci2023GenerativeSC, Nam2023LanguageOrientedCW}. Another similarity is that the semantic communication framework involving DDPMs does not require end-to-end training and we can consider a generic semantic extractor in the transmitter that sends over the channel the semantic information that will be involved to guide the DDPMs generation process.
Diffusion model training is highly stable and generated samples are usually of high quality, although they still suffer from slow sampling.

\subsection{Semantic-Conditioning Generative Models}
\label{sec:conditioning}

\begin{figure}[t]
    \centering
    \includegraphics[width=\linewidth]{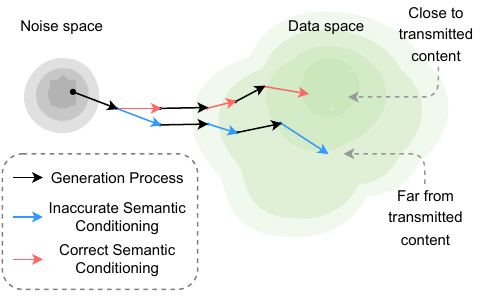}
    \caption{Generative process guided from correct vs inaccurate semantic conditioning. Inaccurate semantic conditioning leads the process far from the transmitted content distribution, while correct semantic conditioning conditions the process towards a generation closer to the original content.}
    \label{fig:gen_process}
\end{figure}

The key element for generative semantic communication frameworks is the semantic conditioning that guides the generation process. In semantic communication, the semantics is often the only information that the receiver has about the original content, therefore a proper and powerful semantic extraction is crucial for content regeneration at the receiver side. As Fig.~\ref{fig:gen_process} shows, since the semantics will guide the direction of the generation process, slightly inaccurate information may lead to a completely incorrect generation far from the transmitted content. Conversely, proper semantic information for conditioning will correctly lead the generation process to the desired content, closer to the transmitted one.

However, extracting the proper semantic representation is not straightforward since there is not a unique recipe to do it yet, as the semantic representation may depend on the type of data or on the task the receiver has to solve. There exist tasks or data for which extracting the semantics is easier, such as segmenting an image to obtain the semantic map, or learning the latent features as in VAEs. Conversely, there are tasks or data types for which getting a semantic representation is tricky, such as multimodal data like video in which one should ideally obtain a representation for both image and audio combined.

Recently, several ways of semantically guiding diffusion models have been proposed, starting from specific-semantic conditioning such as semantic maps \cite{Grassucci2023GenerativeSC} or textual descriptions \cite{ Nam2023LanguageOrientedCW, Nam2023SECON}. Concurrently, more generic approaches started to appear in the literature, allowing multimodal conditioning to generate more and more complex scenes and better align the semantic representations with the desired output \cite{Liu2021MoreCF}.

\section{Evolution of the Communication: Improving Conventional Communication Tasks}
\label{sec:classic}

\subsection{Semantic Compression Beyond Perceptual Compression}

Existing semantic communication transceivers such as deep joint source and channel coding (DeepJSCC) \cite{gunduz2023jsac} aim to remotely reproduce data that are perceptually similar. These methods rely commonly on encoder-decoder structured neural network models such as VAEs. As visualized in Fig.~\ref{fig:taxonomy}, the semantic representations in a VAE are original data statistics $\mu$ and $\sigma$ in a reduced dimension. In computer vision, obtaining these semantic representations is referred to as perceptual compression during which the reproduced outputs maintain perceptual similarities. Under encoder-decoder architectures, any further compression leads to significant perceptual dissimilarities and it is referred to as semantic compression.

Unlike DeepJSCC, generative semantic communication has the potential in enabling semantic compression beyond perceptual compression. For instance, receivers equipped with diffusion models can generate high-fidelity images by feeding random noise. In this case, despite a zero transmit rate, distortion may not always be high. 

Furthermore, sending additional semantic representations and applying semantic conditioning can significantly reduce the distortion, as shown in Fig.~\ref{fig:gen_process}. These additional semantic representations, such as text descriptions and semantic segmentation, are often in lower dimensions than the original data, and therefore, their required rate can be smaller than the rate for $\mu$ and $\sigma$ in perceptual compression.

\subsection{Modular Architecture Beyond End-To-End Architecture}
Deep learning approaches in communication have been successfully employed for joint source-channel coding in an end-to-end framework \cite{gunduz2023jsac}. However, they typically require an overall reformulation of the network architecture. Indeed, conventional schemes need to be able to check whether the received packets contain errors or not before deciding to forward them to the upper layers in the usual OSI stack model.
Conversely, by employing deep generative models, we can still rely on the classical hierarchical OSI model and directly process the content at the application layer. In this way, we still exploit the robustness of current networks, but we add many more degrees of flexibility.

Indeed, differently from any other methods, deep generative models have knowledge of the data distribution and this is the crucial information that can be used to regenerate content even though the received information is corrupted or contains missing data because, for example, of erroneous packet. These models are not required to always receive the whole information as they can exploit their knowledge of the data distribution to consistently generate the missing parts. Similarly, even in the case of highly corrupted received information, deep generative models can denoise it with the help of their intrinsic knowledge of the data.

A class of standard problems in which deep generative models can unveil such capabilities is solving inverse problems. The problem of content transmission over a noisy channel can be modeled as two consequent processes, the forward and inverse problems in which the first one is represented by the sender encoding and the deteriorating process to solve is the transmission with channel corruptions or distortions. Then, the model tries to recover the original data or the key aspects of it from the received noisy and corrupted information at the receiver side. Generative approaches take the corrupted signal in input and directly estimate the clean one without requiring supervision of the corresponding original clean signal. The key aspect is that generative models can regenerate the missing portions or the degraded ones to be consistent with the known parts of the signal thanks to their estimate of the data distribution \cite{Grassucci2023DiffusionMF}. 
This enables non-end-to-end approaches by considering two separate models at receiver and transmitter, increasing the flexibility of communication frameworks whose receivers can interact with multiple sources without requiring joint training.
Several traditional inverse communication tasks can be excellently solved with deep generative models such as denoising, compression, restoration, inpainting, bandwidth extension, and many others.

Furthermore, in addition to improving perceptual results on the data, deep generative models can have a broader impact on the whole network. Communication systems can rely on deep generative models in the case of traffic congestion or problems on the network, as we may transmit less information to avoid congestion and then regenerate the missing one with the generative model avoiding delays and reducing the latency. 

\section{Revolution of the Communication: Towards Emerging Applications}

\label{sec:emerging}


\subsection{Semantic Decomposition for Channel-Adaptive Communication}

Semantic communication may encounter challenges in transmission channels, especially when dealing with fading channels in wireless communication setups. Given the uncertainty of the channel condition before transmission, an adaptive approach becomes crucial. In this context, joint source-channel coding \cite{gunduz2023jsac}, incorporating superposition coding \cite{Tian08}, emerges as an effective strategy to address unknown fading channel conditions. 

\begin{figure}
    \centering
    \includegraphics[width=\linewidth]{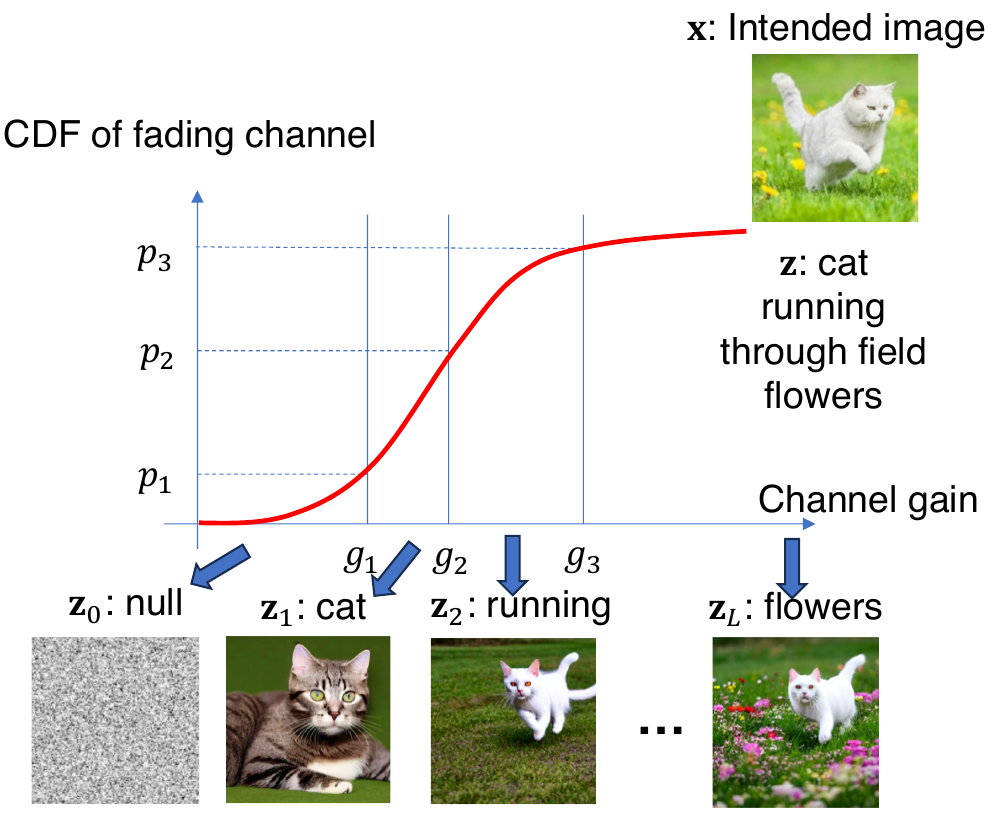}
    \caption{An illustration of superposition coding for semantic coding with generative models.}
    \label{fig:SC_cat}
\end{figure}

Since the channel condition is unknown while no re-transmission is expected due to low latency requirements, it is important to design the generative model for semantic communication to be flexible. In particular, the encoder needs to produce a set of tokens or prompts that allows the decoder to reconstruct the input signal with variable quality. To this end, suppose that the latent variables can be decomposed into a set of $L$ prompts, denoted by $\cZ = \{\bz_1, \ldots, \bz_L\}$, for a given input signal $\bx$. That is, a generative model-based encoder can produce a reconstruction of $\bx$ using a subset of prompts. For convenience, let $D_l = D(\bx, \hat \bx(\cZ_l))$ represent the distortion when the decoder reconstructs $\bx$ with a subset of prompts $\cZ_l = \{\bz_1, \ldots, \bz_l\}$, which is denoted by $\hat \bx(\cZ_l)$. Here, we assume that $\cZ_l$ is more informative than $\cZ_m$ for $m = 1, \ldots, l-1$. Superposition coding can be designed in such a way that the receiver is able to decode more prompts as the channel condition improves. In other words, the receiver can decode none of them or at most $\bz_1$ when the channel condition is poor, while it can decode all the prompts when the channel condition is good. Then, it is expected that $\cD_{l-1} \ge \cD_l$, $l = 1,\ldots, L$. 
In Fig.~\ref{fig:SC_cat}, we illustrate superposition coding for semantic coding with generative models, where the receiver can generate a cat image of different quality depending on the fading channel condition.

As the channel condition improves, the receiver's ability to decode more prompts implies that the quality of the reconstruction of the intended meaning becomes more robust. The adaptive nature of the system ensures that the fidelity of the reconstructed meaning depends on the prevailing channel conditions. This relationship is particularly evident in semantic broadcast communication, where a user with a good channel condition can have a clear understanding, while a user with a poor channel condition may only grasp a basic meaning. For example, if a user can only receive $\bz_1$, he or she can only generate an image of cat, while another user receiving all the prompts can generate an image that is close the intended image.

\subsection{Multimodal Semantic Diversity for Reliable Communication}

Multimodal semantic communication can effectively convey the meaning of a message using various types of data, including video, speech, sketches, and more. In \cite{jiang2023large}, a single semantic encoder is considered for converting all types of data into unimodal text data. This process involves encoding each type of multimodal data with dedicated encoders. The encoded results are then processed by a conditional encoder based on the target modality, which, in this case, is text. The output from the conditional encoder is input into a text diffusion model to generate text data that preserves semantic consistency with the original multimodal data. However, this approach may have a few difficulties. For example, due to the use of conditional encoding, if some of the text prompts are not correctly received, the receiver may not be able to generate multimodal data with sufficient accuracy. To avoid this problem, we can consider multimodal diversity. Each type of data is encoded into unimodal text data, denoted by $\bz_m$, $m = 1,\ldots, M$, where $M$ is the number of types of multimodal data. Then, unlike \cite{jiang2023large}, rather than using conditional encoding sequentially, conditional encoding can be carried out to recover when any $D$ out of $M$ encoded types can be successfully received, while each encoded type is transmitted over an independent fading channel. It is noteworthy that $D$ can be less than $M$ as each type of multimodal data can be correlated with the others. For a large $M$, if $M (1 - p_{\rm err}) > D$, where $p_{\rm err}$ denotes the decoding error probability when transmitting $\bz_m$ over an independent fading channel, we can expect that the full multimodal data can be generated with a high probability thanks to multimodal diversity.



\subsection{Emerging Generative Applications}
\begin{figure}[t]
    \centering
    \includegraphics[width=\linewidth]{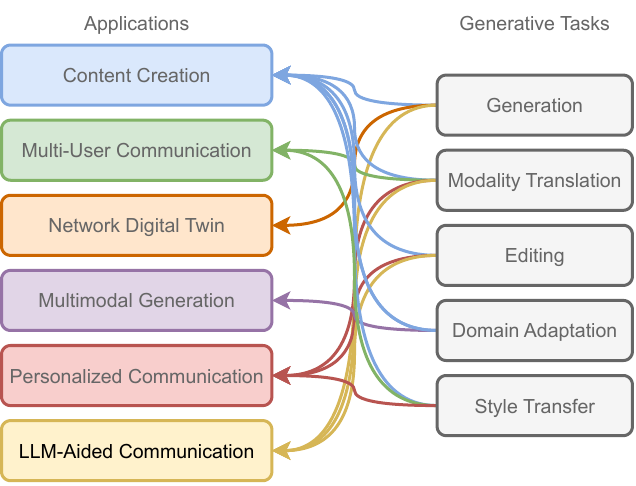}
    \caption{Schematic representation of different possibilities of emerging applications with respective generative tasks that can be accomplished for each application.}
    \label{fig:emerging_app}
\end{figure}

While deep generative models can be involved in classic communication problems, they unleash their real potential when shaping novel applications for semantic communications. Fig.~\ref{fig:emerging_app} shows some emerging applications unveiled from generative models and the corresponding generative tasks that can be accomplished to perform such applications. Besides data generation, the most popular tasks in generative modeling are signal enhancement, style transfer, domain adaptation, cross-modality translation, and content editing, among others. In the following, we see how these tasks can be applied to communication scenarios to give rise to new and emerging applications.

\textbf{Content Creation.}
One of the novelties that generative models allow us to bring to the world of communications is to create information content at the receiver based on the semantics extracted from the information transmitted. This enables the realization of new goal-oriented communication paradigms.
Unlike traditional communication applications, generative models allow the translation of transmitted information into a different representation related to some semantic content. Tasks like signal-to-signal translation or modality translation can be applied to enable such emerging applications. As an example of content creation by modality translation, a text message can be translated into an image rather than into a video or audio signal.

\textbf{Multi-User Communications.}
Multi-user communication may benefit from typical tasks of generative modeling, such as domain adaptation, modality translation rather than style transfer. In such scenarios, different information can be transmitted, each subject to different conditions of the sender but also of the receiver. This often leads to a loss of quality, inhomogeneity and non-uniformity of communication. Based on the encoded semantics, it is possible not only to generate new content, but also to adapt it to the conditions of the end user. This has a significant impact on the quality of communication and resource management.

Furthermore, in such a context, generative models can exploit the combination of semantic spaces related to different users in transmission to provide new content to the end user based on the combined semantics.
As an example, we can think of an image sent by a first user and a text by a second user. The semantics related to the text can be used to edit the image sent by the first user. As a result the end user directly receives the modified information based on the combined semantics.

\begin{figure}[t]
    \centering
    \includegraphics[width=\linewidth]{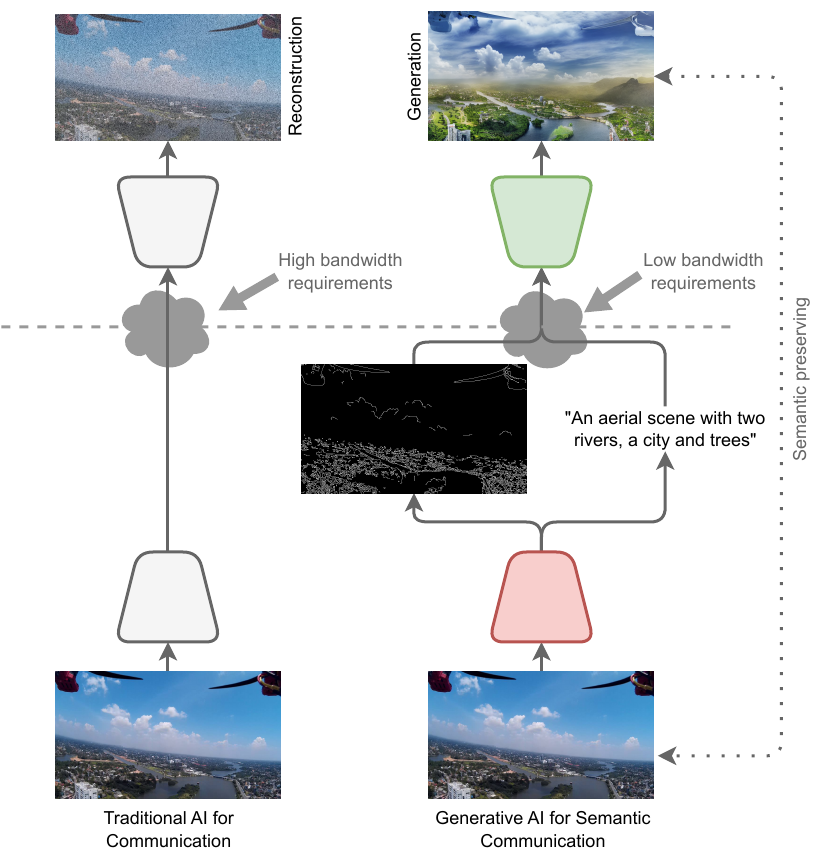}
    \caption{Traditional AI communication scheme and novel generative AI semantic communication architecture. The latter ensures lower bandwidth requirements by transmitting one or more semantic representations (canny edges and text) while guaranteeing message semantic preservation at the receiver. For the semantic extractor and the generative model in the generative AI semantic scenario, we use ControlNet.}
    \label{fig:ai_gsc}
\end{figure}

\textbf{Network Digital Twin.} 
Network digital twin refers to the virtual representation of physical network systems that can be used to study the behavior of specific network systems. The semantic information in this case can be related not only to the information content but also to the network system condition. Thus, if the network model changes due to some different physical network condition, a semantic-guided generative model can adapt the generated content to the new domain of the network representation, without affecting the quality of the communication.

\textbf{Multimodal Generation.}
Since semantic communication is no longer depends on data as traditional communications but rather on its semantic representation, we may think of sending a certain type of signal while receiving it endowed with additional modalities. A typical generative modeling task in that case is represented by cross-modality translation. One application example may be the \textit{semantic-aware sonorization}, where silent video can be received accompanied by music generated based on the semantics of the video. Another application example is represented by \textit{virtual/augmented reality} where generative models can create 3D visual-audio renders based on the semantics extracted from a text rather than an image or audio signal.

\textbf{Personalized Communication.}
Style transfer, along with the other possible tasks of generative modeling, can be adopted for one of the most fascinating applications of future communications, that is the personalized communication. The transmitted information could be personalized based on some semantics representing the user taste rather than the subjective quality perceived by a single user. Generative models can adapt the semantic information to the style desired by the end user.

\textbf{LLM-Aided Communication.}
Large language models (LLMs) are envisaged to be integrated in future communication systems to leverage their unique capabilities of reasoning and in-context learning \cite{lin2023pushing,Nam2023LanguageOrientedCW}. Generative models can play a crucial role in translating various data modalities into natural language compatible with LLMs. Moreover, they can also transform the text outputs of LLMs into intended data modalities, maximizing their effectiveness in specific tasks.

\section{Generative AI Challenges and Future Perspectives}
\label{S:Challenges}
Despite the exciting results and the novel tasks and applications that deep generative models are unveiling, the new challenges such models are bringing have yet to be completely overcome.

\textbf{Sustainability.} The first challenge to face surely regards computational efficiency and sustainability. Deep generative models require a huge amount of resources in terms of computational power, storage memory, and time both for training and inference, caused by the computations on billions of parameters. Nevertheless, there exist some pathways to undertake for reducing the impact of deep generative models on computational resources. Promising results are coming from the extremely low-bit quantization of diffusion and large language models, in which networks can be defined in 4 or even 2 bits, with consequent acceleration of both training and inference process up to 10x, as well as a crucial reduction of required runtime and storage memory \cite{li2023qdm}. Merging low-bit quantizations and energy management methods may lead to novel models specifically tailored for communication frameworks that preserve the performance of their large-scale counterparts while being more efficient.

\textbf{Trustworthiness.} The second challenge of deep generative models for semantic communication is their reliability and trustworthiness. 
Indeed, a critical concern for employing these models in communication scenarios is ensuring their robustness against attacks from malicious agents. In these cases, the model may assume an unexpected behavior, and their outputs may no longer be reliable. Similarly, studying privacy leakages in deep generative models is emerging as a noteworthy research topic, as generative models may sometimes be guided to reveal some private information used for the training process. Recently, promising approaches have been presented to address both these concerns: training the model to be robust to adversarial attacks, and avoiding training over the same data may effectively alleviate data leakage at inference time. 

\textbf{Explainability.} Finally, a third challenge to solve regards the explanations and the interpretations we can give to the generated samples of generative models and to their behavior. Indeed, in communication scenarios, we should be able to explain a given output and what happens inside the model to produce that sample in order to guarantee the quality and reliability of the generation. Moreover, explaining the generation process and demystifying the ``black box" belief around deep networks may increase the confidence of users in these methods and spread their adoption.

\section{Conclusions}   \label{S:Conc}
Deep generative models have suddenly entered our daily lives and are revolutionizing several fields of application. In this paper, we present the advantages of employing them to solve classic communication inverse problems, nevertheless, since deep generative models unveil their real potential under the new paradigm of semantic communication, we expound on the novel possibilities, tasks, and applications these models are bringing to semantic communication frameworks. Finally, we discuss the new challenges such methods are posing and delineate some possible paths to solve them, especially from the efficiency point of view.

\section*{Acknowledgment}
This work was partially supported by the European Union under the Italian National Recovery and Resilience Plan (PNRR) of NextGenerationEU, partnership on
``Telecommunications of the Future” (PE00000001 - program RESTART), and in part by the MSIT (Ministry of Science and ICT), Korea, under the ITRC (Information Technology Research Center) support program (IITP-2023-RS-2023-00259991) supervised by the IITP (Institute for Information \& Communications Technology Planning \& Evaluation).

\bibliographystyle{ieeetr}
\bibliography{si}

\end{document}